\documentclass[10pt,letterpaper]{article}

\usepackage{arxiv}
\usepackage[T1]{fontenc}
\usepackage[utf8]{inputenc}
\usepackage{microtype}
\usepackage{amsmath}
\usepackage{amssymb}
\usepackage{graphicx}
\usepackage{booktabs}
\usepackage{multirow}
\usepackage{cite}
\usepackage{algorithm}
\usepackage{algorithmic}
\usepackage{placeins}
\usepackage{float}
\usepackage{colortbl}
\usepackage[hidelinks]{hyperref}

\makeatletter
\renewcommand\floatc@ruled[2]{{\@fs@cfont #1:} #2\par}
\makeatother

\definecolor{BestYellow}{RGB}{255,243,178}
\definecolor{WorstBlue}{RGB}{213,231,255}

\renewcommand{\headeright}{}
\renewcommand{\undertitle}{}
\renewcommand{\shorttitle}{VANE}

\title{VANE: Reliable Test-Time Training for Vision-Language-Action Models via Future Visual Representation Prediction}

\author{%
\sffamily\small
Hongjin Ji\textsuperscript{1,5}\thanks{Equal contribution.\quad
\textsuperscript{\ensuremath{\dagger}}Project leader.\quad
\textsuperscript{\ensuremath{\ddagger}}Corresponding author.\\
Email: \href{mailto:hongjinji@link.cuhk.edu.cn}{\texttt{hongjinji@link.cuhk.edu.cn}}}\quad
Guoyang Xia\textsuperscript{2,5,*}\quad
Luoyang Sun\textsuperscript{3,4,5,*}\quad
Fangxiang Feng\textsuperscript{2}\quad
Lei Ren\textsuperscript{5,\ensuremath{\dagger},\ensuremath{\ddagger}}\\[0.55em]
\sffamily\mdseries\footnotesize
\textsuperscript{1}The Chinese University of Hong Kong, Shenzhen\quad
\textsuperscript{2}Beijing University of Posts and Telecommunications\\
\textsuperscript{3}Institute of Automation, Chinese Academy of Sciences\\
\textsuperscript{4}University of Chinese Academy of Sciences\\
\textsuperscript{5}Li Auto Inc.
}
\date{\vspace{-0.4in}}

\begin{document}

\maketitle

\begin{abstract}
Test-time training (TTT) offers a lightweight way to adapt
vision--language--action (VLA) policies from unlabeled deployment streams, but
it remains difficult to use reliably in closed-loop manipulation. A shared
adaptation space can mix incompatible task corrections, while an online update
can alter subsequent actions before its consequences are known. We introduce a reliable TTT framework for VLA policies ~(\emph{VANE}). VANE conditions prompt
adaptation on the current vision--language context and learns from the future
visual consequences of executed actions. Candidate updates are isolated from
the live policy, evaluated on subsequent observations, and committed only when
supported by future evidence, making adaptation selective and reversible. On
SimplerEnv WidowX, VANE improves average success by $3.2$ percentage points over
the corresponding TTT baseline. Results on Google Robot further show that
deployment-time gains remain task- and embodiment-dependent. Together, these
results demonstrate a constrained, evidence-based approach to adapting VLA
policies during interaction.
\end{abstract}

\section{INTRODUCTION}
\label{sec:introduction}
An expanding family of vision--language--action (VLA) models provides strong
policy backbones for robot manipulation
\cite{rt2,openvla,pi0,starVLA-alpha}. Nevertheless, adapting these backbones to
new tasks or visual environments remains expensive even when the robot
embodiment is fixed. Conventional adaptation commonly requires additional
task-specific demonstrations with action labels, followed by repeated
fine-tuning and hyperparameter selection. As the number of downstream tasks
grows, this process repeatedly consumes data and engineering effort and often
produces a collection of separately tuned policies rather than one maintainable
deployed system. In contrast, robot deployment naturally produces streams of
unlabeled visual observations. A practical adaptation mechanism should exploit
this readily available supervision, preserve a shared VLA backbone, represent
task-dependent corrections, and prevent harmful updates from directly entering
the closed-loop controller.

Test-time training (TTT) offers a promising alternative by optimizing a
self-supervised objective on unlabeled test observations while updating only a
small parameter subset \cite{ttt}. TTT-VLA \cite{ttt-vla} further demonstrates
that a latent prompt can connect a self-supervised proxy task to a frozen VLA
policy, avoiding full-model fine-tuning during deployment. This establishes the
feasibility of lightweight VLA TTT, but does not yet answer whether the resulting
updates remain dependable across heterogeneous tasks and closed-loop
interaction. In our controlled study on QwenPi
\cite{starVLA-alpha}, latent-prompt TTT produces inconsistent outcomes across
tasks and visual conditions: an update that benefits one setting may degrade
another, while an unvalidated online update may reduce the performance of the
deployed policy. In this work, we call TTT \emph{reliable} when candidate
updates are introduced selectively and regressions caused by task interference
or erroneous online optimization are suppressed. We identify three design
challenges to achieving this behavior.

First, prompt adaptation is strongly task-dependent. At a fixed checkpoint,
the non-adapted policy obtains a success rate of $64.1\%$.
Optimizing one prompt jointly over all four tasks decreases the result to
$63.3\%$, whereas independently adapted task-specific prompts reach $66.1\%$.
The gap indicates that a single shared prompt entangles heterogeneous, and
sometimes conflicting, correction directions. At the same time, fully
independent prompts discard potentially reusable structure across related
tasks. We therefore seek an adaptation space that preserves sharing while
allowing the current vision--language context to compose different corrections.
This motivates a context-routed \emph{Mixture of Latent Prompts} (MoLP), which
constructs the effective prompt from a shared latent-prompt bank without
requiring an externally supplied task identifier. We analyze the underlying
task-sharing interference in \autoref{sec:task_interference_analysis} and
evaluate the resulting prompt structure in
\autoref{sec:controlled_evaluation}.

Second, the usefulness of prompt-only TTT depends critically on its proxy
objective. A suitable deployment proxy should satisfy two requirements. First,
its target must be obtainable from ordinary robot interaction during both
training and deployment, without expert actions, rewards, success labels, or
privileged evaluation signals. Second, optimizing it should provide a
policy-relevant adaptation signal: the target should preserve semantic
structure that can generalize across tasks and visual conditions while exposing
interaction-dependent change rather than merely reconstructing what is already
visible. State grounding satisfies the availability requirement because the
current robot state is directly observed, but its target is low-dimensional,
embodiment-specific, and synchronized with the current image. It can therefore
be optimized without explicitly explaining how the visual scene evolves after
interaction. Delayed RGB observations are equally label-free, and a frozen
video representation encoder converts them into semantically structured future
visual targets without requiring pixel-level generation. Predicting such a
target from the current observation, adaptable prompt, and action context yields
a temporally predictive and action-conditioned proxy. We call this supervision
branch the \emph{World-Predictive Interface} (WPI).

We instantiate this proxy design on \emph{QwenPi}, a Qwen-based flow-matching
VLA constructed within the StarVLA-$\alpha$ framework, rather than on the native
$\pi_{0.5}$ policy used by TTT-VLA. We retain the QwenPi action-policy backbone
and replace its state-grounding proxy branch with WPI; the resulting variant is
called \emph{QwenWPI}. Here, \emph{world-predictive} denotes a
world-model-like supervision interface that predicts future visual
representations. QwenWPI is not a standalone world model and does not replace
the action policy. Matched comparisons between state grounding and WPI are
reported in \autoref{sec:controlled_evaluation}.

Third, replacing a same-time state proxy with future-representation prediction
changes the online optimization problem itself. With state-grounded QwenPi, a
low-dimensional proxy target is available at the current step and a prompt
update can be formed immediately. With QwenWPI, a valid objective becomes
available only after the delayed observation $o_{t+k}$ arrives, and each pair
requires encoding and comparing token-level visual representations. More
importantly, immediately deploying a candidate prompt changes the actions that
generate its subsequent supervision: the update can alter the very future
observations on which it would later be judged. Reliable predictive TTT must
therefore decide not only when an update is informative, but also how to
evaluate it without allowing the candidate to control the trajectory used for
validation.

We address this difficulty with \emph{Attention-Gated and Validation-Driven
Test-Time Training} (AGV-TTT). Cross-layer attention redistribution provides a
policy-internal signal for proposing a sparse update near interaction
transitions. The update is applied to a shadow copy of the latent-prompt bank,
while the live prompt continues to control the robot. As subsequent observations
arrive, the old and candidate prompts are evaluated on matched future-prediction
pairs. The candidate prompt and its optimizer state are committed atomically
only when the focused objective improves without global regression; otherwise,
the candidate is discarded. Attention therefore determines \emph{when to
propose} adaptation, while delayed observations determine \emph{whether the
proposal should be deployed}. The attention pattern motivating the proposal
cue is examined in \autoref{sec:attention_event}, and the contribution of
future validation is isolated in \autoref{sec:validation_ablation}.

These considerations lead to \emph{VANE}, which combines a context-routed
latent-prompt adaptation space, a future-predictive proxy, and a selective
proposal--validation--commit protocol. The name VANE evokes a weather vane, 
which responds to changes in its surroundings while remaining anchored to 
a stable structure, reflecting our goal of selectively adapting a shared policy to 
changing deployment conditions. MoLP represents task-dependent corrections, 
QwenWPI supplies observable and policy-relevant self-supervision, and AGV-TTT 
handles the delayed and closed-loop nature of predictive TTT. At deployment, 
all policy, routing, and predictive modules remain frozen; only the 
latent-prompt bank is eligible for validated updates.

Our contributions are summarized as follows:
\begingroup
\setlength{\leftmargini}{2.5em}
\begin{itemize}
    \item We systematically diagnose latent-prompt TTT on QwenPi and show that
    heterogeneous manipulation tasks require context-dependent corrections. We
    propose a context-routed Mixture of Latent Prompts (MoLP) that balances
    cross-task sharing and specialization within a compact adaptation space.

    \item We formulate two requirements for a deployment-time proxy---target
    observability and policy relevance---and instantiate future visual
    representation prediction as the World-Predictive Interface. Replacing the
    state-grounding branch of QwenPi with WPI yields QwenWPI, which uses delayed
    observations as a temporally predictive, label-free objective for
    prompt-only TTT.

    \item We expose two challenges introduced by predictive TTT---delayed
    targets and closed-loop coupling---and develop AGV-TTT to address them.
    AGV-TTT converts
    policy-internal attention transitions into isolated shadow proposals, then
    validates each candidate on matched future observations before atomic commit
    or rollback.
\end{itemize}
\endgroup
\FloatBarrier

\begin{figure}[H]
    \centering
    \includegraphics[width=\textwidth]{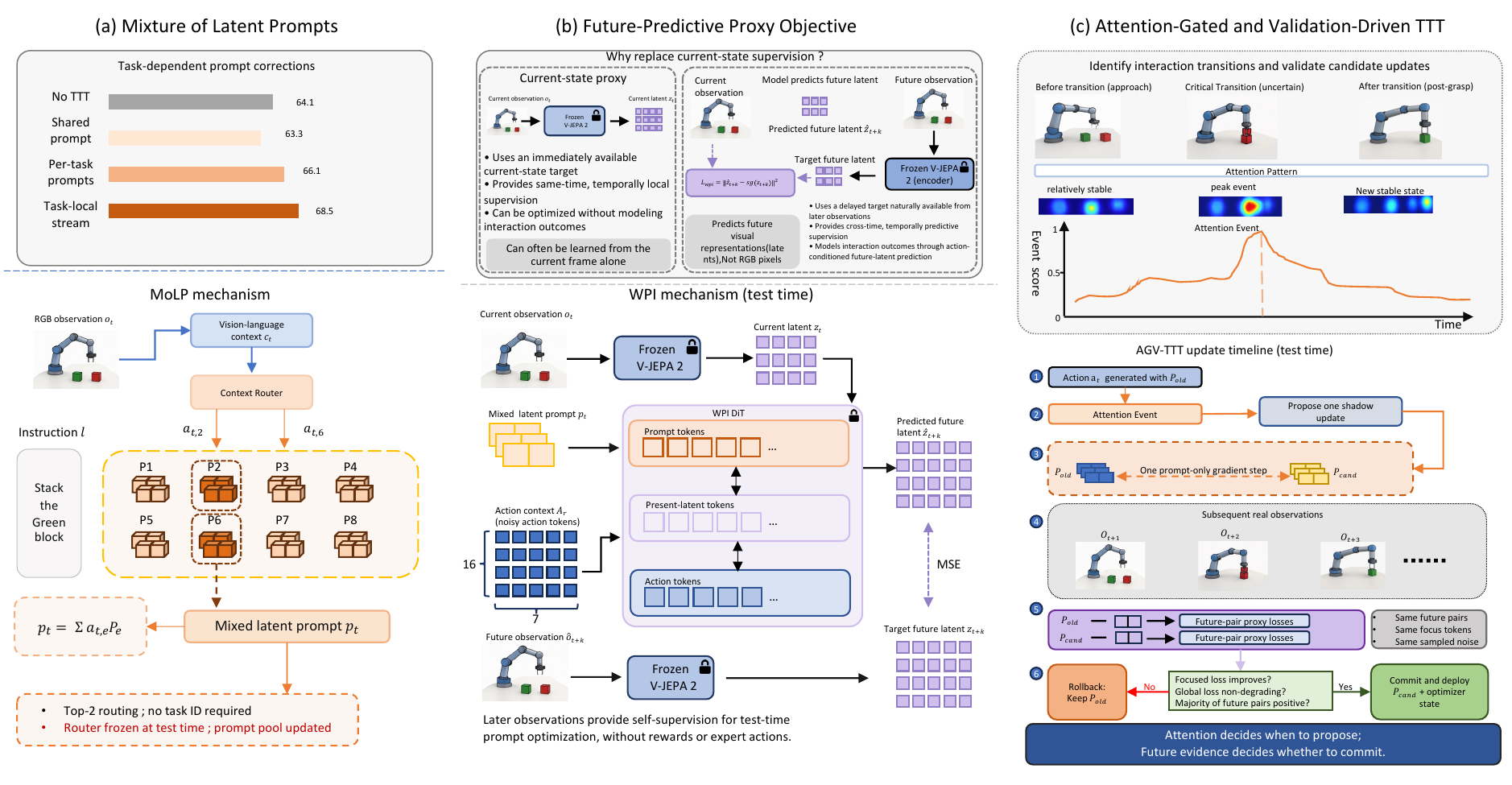}
    \caption{Overview of VANE. (a) MoLP composes a context-dependent latent
    prompt from a shared prompt bank without requiring a task identifier.
    (b) The upper comparison contrasts two deployment-observable proxies:
    state grounding supplies an immediately available but same-time target,
    whereas WPI uses delayed observations to provide cross-time,
    action-conditioned supervision of interaction outcomes. The lower diagram
    instantiates WPI as future visual representation prediction for prompt-only
    TTT. (c) AGV-TTT uses cross-layer attention redistribution to propose a
    shadow update and commits it only after paired future validation.}
    \label{fig:vane_overview}
\end{figure}

\section{RELATED WORK}

\subsection{VLA Policies and Predictive Interfaces}
Generalist vision--language--action policies learn transferable manipulation
skills from large-scale robot data and vision--language pretraining
\cite{rt1,rt2,openx,octo,openvla}. Recent policies such as $\pi_0$ and
StarVLA-$\alpha$ formulate continuous action-chunk generation through flow
matching \cite{pi0,starVLA-alpha}. Our experiments use QwenPi, a Qwen-based flow-matching VLA instantiated
following StarVLA-$\alpha$, as the controlled action-policy backbone. QwenWPI
preserves the same VLM and action-policy modules and changes only the proxy
interface, isolating the effect of future visual latent prediction.

Predictive representation learning provides supervision by modeling how visual
observations evolve. V-JEPA~2 predicts future video representations in latent
space, while predictive VLAs incorporate future dynamics into policy
pretraining or joint action learning \cite{vjepa2,futurevla,lingbot-va}.
QwenWPI instead keeps the visual target encoder and policy backbone frozen and
uses future visual representations as a deployment-time objective for
prompt-only optimization.

\subsection{Test-Time Training for LLMs, VLMs, and VLA Policies}
Test-time learning began in vision with self-supervised training or
entropy-based adaptation on unlabeled test samples
\cite{ttt,tent,cotta,eata}. The idea later reached foundation models: LLMs can
train on retrieved neighboring text before inference \cite{ttt-llm}, while
VLMs can tune prompts per image using consistent, confident predictions across
augmented views \cite{tpt}. These studies establish label-free adaptation from
test inputs, but focus on static prediction, where an update does not determine
the next observation.

VLAs add a closed-loop challenge: an update changes subsequent actions and thus
the data available for further adaptation. TTT-VLA jointly trains action
prediction and state grounding, then adapts only latent prompts at deployment
\cite{ttt-vla}. Concurrently, $T^3$VF uses predicted--attained
future-image pairs and action-variance filtering to adapt visual-foresight
policies \cite{t3vf}, while reward-based methods require an explicit
test-time reward signal \cite{tt-vla}. VANE retains latent-prompt adaptation
but adds context routing, future-latent supervision, and future validation of
shadow updates before closed-loop deployment.

\section{METHOD}
\label{sec:method}
\subsection{Overview}
\label{sec:method_overview}

Given an RGB observation $o_t$ and language instruction $\ell$, the QwenPi
policy predicts an action chunk $\widehat A_t\in\mathbb R^{16\times7}$ while
being conditioned on an effective latent prompt $p_t$:
\begin{equation}
    \widehat A_t=\pi_{\Theta}(o_t,\ell;p_t).
    \label{eq:policy_compact}
\end{equation}

Figure~\ref{fig:vane_overview} presents VANE as a sequential adaptation
pipeline rather than three independent modules: MoLP determines \emph{what} to
adapt, WPI supplies the adaptation signal, and AGV-TTT decides \emph{when} to
propose an update and \emph{whether} to deploy it.

First, MoLP sparsely routes the current vision--language context over a shared
prompt bank to form $p_t$, without requiring a task identifier. This mitigates
cross-task interference and confines adaptation to a structured prompt space.
WPI then conditions on $p_t$, the current visual latent, and action tokens to
predict a future visual latent, using later observations as label-free targets.
Unlike a current-state proxy, it models action-induced future change without
rewards, expert actions, or success labels.

AGV-TTT governs the resulting online update. The deployed prompt
$P_{\mathrm{old}}$ acts until an attention transition triggers one prompt-only
step on a shadow copy, producing $P_{\mathrm{cand}}$ while
$P_{\mathrm{old}}$ remains live. Subsequent observations evaluate both prompts
with matched future pairs and controlled inputs. The candidate is committed
only if its focused loss improves, its global loss does not degrade, and most
future pairs favor it; otherwise, the update is rolled back. Thus, attention
decides when to propose, while future evidence decides whether to commit.

During robot-data training, the policy, latent-prompt bank, and router are
learned jointly with
\begin{equation}
    \mathcal{L}_{\mathrm{train}}
    =\mathcal{L}_{\mathrm{act}}
    +\lambda_{\mathrm{wpi}}\mathcal{L}_{\mathrm{wpi}}
    +\lambda_{\mathrm{lb}}\mathcal{L}_{\mathrm{lb}},
    \label{eq:training_compact}
\end{equation}
where $\mathcal{L}_{\mathrm{act}}$ is the original flow-matching action loss,
$\mathcal{L}_{\mathrm{wpi}}$ is the future visual representation loss, and
$\mathcal{L}_{\mathrm{lb}}$ regularizes routing utilization. At deployment,
all modules except the latent-prompt bank are frozen, yielding a constrained,
verifiable, and reversible adaptation process.
For reference, the notation used throughout the method is summarized in
Appendix~\ref{app:notation}, and implementation details are provided in
Appendix~\ref{app:training_details}.
\subsection{Context-Routed Mixture of Latent Prompts}
\label{sec:latent_prompt_molp}

A single latent prompt forces heterogeneous task corrections into one shared
parameter block, whereas fully independent prompts discard useful cross-task
structure. We instead maintain a global latent-prompt bank
$\mathcal{P}=\{P_i\}_{i=1}^{E}$ and learn a sparse, context-dependent
composition. Let $c_t$ be the last valid token of the final projected VLM
representation, which jointly summarizes the current image and instruction.
A linear router produces prompt-selection probabilities and the effective
prompt:
\begin{equation}
\begin{aligned}
    q_t &= \operatorname{softmax}(W_r c_t+b_r),\\
    \mathcal{S}_t &= \operatorname{TopK}(q_t,K),\\
    p_t &= \sum_{i\in\mathcal{S}_t}
    \frac{q_{t,i}}{\sum_{j\in\mathcal{S}_t}q_{t,j}}P_i.
\end{aligned}
    \label{eq:molp_compact}
\end{equation}
We use a shared bank of 8 latent prompts with top-2 routing. The bank is not
partitioned by task or embodiment, allowing related tasks to reuse prompt
components while different vision--language contexts compose distinct prompt
mixtures. The routing weights are recomputed for each new control observation
and then held fixed throughout the flow-matching iterations used to generate
the corresponding action chunk.

The router and latent prompts are trained jointly with the action and proxy
objectives, together with a standard load-balancing loss. During test-time
training, the router is frozen and only the latent-prompt bank $\mathcal{P}$ is
optimized. This defines a compact adaptation space whose behavior is conditioned
on the current vision--language context without requiring an externally
provided task identifier. Initialization, load balancing, and optimizer details
are provided in Appendix~\ref{app:training_details}.
\subsection{World-Predictive Interface}
\label{sec:wpi}

A deployment-time proxy must balance target observability with policy
relevance. State grounding supplies an immediately available current-state
target, but provides same-time, embodiment-specific supervision and can be
optimized without modeling the outcome of an interaction. WPI preserves
label-free target availability by using later observations and asks the
adaptable prompt to predict their frozen visual representations from the
current observation and action context. It therefore replaces temporally local
state supervision with cross-time, action-conditioned supervision while
retaining the original QwenPi action-policy backbone.

A frozen V-JEPA 2 encoder $f_{\omega}$ extracts present and delayed future
visual tokens, $Z_t=f_{\omega}(o_t)$ and
$Z_{t+k}=f_{\omega}(o_{t+k})$. The present tokens, routed prompt, and noised
action tokens form the Latent-Action DiT query
\begin{equation}
    Q_{\tau}^{(0)}=
    \left[p_t\;\middle|\;E_z(Z_t)\;\middle|\;E_a(A_{\tau},\tau)\right].
    \label{eq:wpi_query_compact}
\end{equation}
Visual-latent and action tokens exchange information through shared
self-attention, so the prediction is conditioned on the policy's action-related
context rather than on the current image alone. The slots initialized from
$Z_t$ are decoded directly into the delayed representation
$\widehat Z_{t+k}$ and trained with
\begin{equation}
    \mathcal{L}_{\mathrm{wpi}}
    =\frac{1}{NC}
    \left\|\widehat Z_{t+k}-\operatorname{sg}(Z_{t+k})\right\|_F^2,
    \label{eq:wpi_loss_compact}
\end{equation}
where $N$ and $C$ are the number and width of the visual tokens. During
training, this objective is optimized jointly with action generation. During
deployment, the delayed observation supplies the target without rewards,
success labels, or expert actions, and gradients are applied only to the
latent-prompt bank. The exact flow-matching objective, prediction horizon, and
attention-mask implementation are deferred to
Appendix~\ref{app:training_details}.

\subsection{Attention-Gated and Validation-Driven TTT}
\label{sec:agv_ttt}

Online TTT treats every valid proxy pair as equally suitable for an
update. Our attention visualizations reveal a clear temporal structure:
routine motion occupies a comparatively stable attention regime, whereas
contact, grasp closure, and lift-off induce pronounced cross-layer
redistribution. AGV-TTT uses this policy-internal transition signal to decide
when an update is worth attempting, while reserving future observations to
decide whether the attempted update should be deployed.

\paragraph{Endogenous event detection.}
The live prompt first generates $a_t$ under normal policy inference. During
that forward pass, we collect (i) action-to-VLM cross-attention and (ii)
action-to-predictive-latent attention. For channel $c$, denoising step $u$, and
DiT layer $l$, let $a^c_{t,u,l}$ be the normalized attention distribution. We
aggregate its temporal variation as
\begin{equation}
    \Delta_t^c=
    \frac{1}{|\mathcal U||\mathcal L|}
    \sum_{u,l}\frac{1}{2}
    \left\|a^c_{t,u,l}-a^c_{t-1,u,l}\right\|_1.
    \label{eq:event_change_compact}
\end{equation}
An event is triggered when either channel is anomalous relative to its recent
robust baseline. Because detection occurs after $a_t$ is generated, a proposal
created by this event cannot retroactively influence the action that triggered
it. The robust normalization and thresholds are specified in
Appendix~\ref{app:agv_details}.

\paragraph{Attention-focused shadow proposal.}
At an event, the predictive-latent attention selects a set
$\mathcal K_t$ of high-attention visual tokens and assigns normalized
weights $F_{t,n}$. With token-wise future-prediction error $e_{t,n}$, the
proposal loss is
\begin{equation}
\begin{aligned}
    \mathcal{L}_{\mathrm{focus},t}
    &=\sum_{n\in\mathcal K_t}F_{t,n}e_{t,n},\\
    \mathcal{P}^{\mathrm{cand}}
    &=\operatorname{Update}
      (\mathcal{P}^{\mathrm{old}};\mathcal{L}_{\mathrm{focus},t}).
\end{aligned}
    \label{eq:focus_proposal_compact}
\end{equation}
The selected tokens and weights are detached and frozen for the entire
proposal-validation cycle. The update is executed on a shadow copy; the live
prompt and optimizer state are immediately restored, so the candidate cannot
affect control before validation.

\paragraph{Future validation and atomic deployment.}
Over the next $H_v$ valid WPI pairs, the old and candidate prompts are evaluated
with identical inputs, focus weights, and flow-matching randomness. Let
$R_{\mathrm{focus}}$ and $R_{\mathrm{global}}$ denote their relative
improvements in the focused and all-token prediction losses, respectively. A
candidate is accepted only when
\begin{equation}
    \begin{gathered}
        R_{\mathrm{focus}} > 0,
        \qquad
        R_{\mathrm{global}} \geq 0,
        \\[2pt]
        \sum_{j=1}^{H_v}
        \mathbb{I}\!\left[
            R_{\mathrm{focus}}^{(j)} > 0
        \right]
        >
        \frac{H_v}{2}.
    \end{gathered}
    \label{eq:acceptance_compact}
\end{equation}
Thus, an update must improve the event-focused representation, avoid global
regression, and remain beneficial on a temporal majority of subsequent pairs.
If accepted, the complete prompt bank and its optimizer state are committed
atomically; otherwise, both are discarded. The first $H_v-1$ validation-frame
actions always use the old prompt, while the final validation is resolved before
that frame's action is generated. This preserves the causal separation between
proposal data, validation evidence, and deployment.
Appendix~\ref{app:agv_details} provides the paired-randomness construction, the
state-grounded AGV instance, and the full deployment procedure summarized in
Algorithm~\ref{alg:wpi_agv_ttt}.
\section{ANALYSIS AND EXPERIMENTS}
\label{sec:analysis_experiments}

We separate two kinds of evidence. The \emph{Diagnostic and Mechanistic
Analysis} subsection examines phenomena that motivate or contextualize
VANE---task-dependent transfer under shared prompts, policy-internal attention
redistribution, and visual-representation shift across evaluation suites. These
observations are not treated as direct proof of method efficacy. The
\emph{Experimental Evaluation} subsection then evaluates the resulting design
through benchmark comparisons, controlled component comparisons, a validation
ablation, and optimization-cost measurements.

\subsection{Diagnostic and Mechanistic Analysis}
\label{sec:analysis}

\subsubsection{Task-Dependent Prompt Interference}
\label{sec:task_interference_analysis}
We revisit the single-prompt assumption using a fixed checkpoint of the
state-grounded policy. Table~\ref{tab:task_sharing} reports success-rate changes relative
to No TTT when a prompt is adapted on different task subsets and then evaluated
across all four tasks. The task-specific row uses a separately adapted prompt
for each evaluation task; every other row uses one prompt jointly optimized on
the named subset. Yellow and blue cells mark the largest gain and degradation
in each column, respectively.

\begin{table}[H]
    \centering
    \caption{Task-sharing diagnosis for state-grounded single-prompt Offline TTT on WidowX at a fixed checkpoint.}
    \label{tab:task_sharing}
    \small
    \setlength{\tabcolsep}{9pt}
    \begin{tabular}{lccccc}
        \toprule
        Adaptation subset & Stack & Carrot & Spoon & Eggplant & Avg. \\
        \midrule
        \multicolumn{6}{l}{\emph{One-task adaptation}} \\[1pt]
        Task-specific prompts
        & $+2.5$ & $+3.1$ & $0.0$ & $+2.6$ & $+2.0$ \\
        \addlinespace[6pt]
        \multicolumn{6}{l}{\emph{Two-task shared prompt}} \\[1pt]
        Carrot + Eggplant
        & \cellcolor{BestYellow}\textbf{$+5.2$} & $+10.4$ & $+1.1$ & $0.0$
        & \cellcolor{BestYellow}\textbf{$+4.1$} \\
        Carrot + Spoon
        & $-1.1$ & $+4.2$ & $+2.1$ & $+1.0$ & $+1.5$ \\
        Spoon + Eggplant
        & $+4.1$ & $+4.2$ & $-2.1$
        & \cellcolor{WorstBlue}\textbf{$-1.1$} & $+1.3$ \\
        Stack + Carrot
        & \cellcolor{WorstBlue}\textbf{$-2.1$} & $+1.0$ & $+1.1$ & $+1.0$ & $+0.2$ \\
        Stack + Eggplant
        & $+4.1$ & $+11.5$ & \cellcolor{WorstBlue}\textbf{$-6.2$} & $+1.0$ & $+2.6$ \\
        Stack + Spoon
        & $+4.1$ & $+8.3$ & $0.0$ & \cellcolor{BestYellow}\textbf{$+3.1$} & $+3.9$ \\
        \addlinespace[6pt]
        \multicolumn{6}{l}{\emph{Three-task shared prompt}} \\[1pt]
        Carrot + Spoon + Eggplant
        & \cellcolor{WorstBlue}\textbf{$-2.1$}
        & \cellcolor{BestYellow}\textbf{$+16.7$}
        & $-2.1$
        & \cellcolor{BestYellow}\textbf{$+3.1$}
        & \cellcolor{BestYellow}\textbf{$+4.1$} \\
        Stack + Carrot + Eggplant
        & $+2.0$ & $+5.2$ & $0.0$ & \cellcolor{BestYellow}\textbf{$+3.1$} & $+2.6$ \\
        Stack + Carrot + Spoon
        & $+3.1$ & $+2.1$ & \cellcolor{BestYellow}\textbf{$+4.2$} & $+2.1$ & $+2.8$ \\
        Stack + Spoon + Eggplant
        & $+2.0$ & \cellcolor{WorstBlue}\textbf{$-1.1$} & $0.0$ & $+1.0$ & $+0.5$ \\
        \addlinespace[6pt]
        \multicolumn{6}{l}{\emph{All-task shared prompt}} \\[1pt]
        Stack + Carrot + Spoon + Eggplant
        & $-0.6$ & $+2.1$ & $-5.7$ & $+1.0$
        & \cellcolor{WorstBlue}\textbf{$-0.8$} \\
        \bottomrule
    \end{tabular}
\end{table}

\paragraph{Observation.}
\textbf{Prompt adaptation transfers unevenly across tasks, and a single
all-task prompt can underperform both the non-adapted and task-specific
alternatives.}
Using a separately adapted prompt for each evaluation task raises the composite
four-task average from $64.1\%$ to $66.1\%$, whereas jointly adapting one
shared prompt on all four tasks lowers it to $63.3\%$. Intermediate subsets
show strongly nonuniform transfer: Carrot + Eggplant and Carrot + Spoon +
Eggplant tie for the largest average gain of $4.1$ points, while several other
subsets improve some evaluation tasks but degrade others. In particular,
all-task adaptation reduces Spoon performance by $5.7$ points. These
outcome-level transfer patterns are consistent with task-dependent interference
when heterogeneous contexts share one adapted prompt. They motivate a
context-routed latent-prompt bank, while the efficacy of MoLP itself is
evaluated separately in Sec.~\ref{sec:controlled_evaluation}.

\subsubsection{Action-to-VLM Attention Changes at Interaction Transitions}
\label{sec:attention_event}
We analyze action-to-VLM cross-attention, which captures how action queries
attend to VLM image-patch tokens during a rollout. We track changes in this
attention distribution across DiT layers and over time. In QwenWPI, AGV
additionally monitors action-to-predictive-latent attention during deployment,
but that additional channel is not required for the diagnostic below.

\begin{figure}[H]
    \centering
    \includegraphics[width=0.98\textwidth]{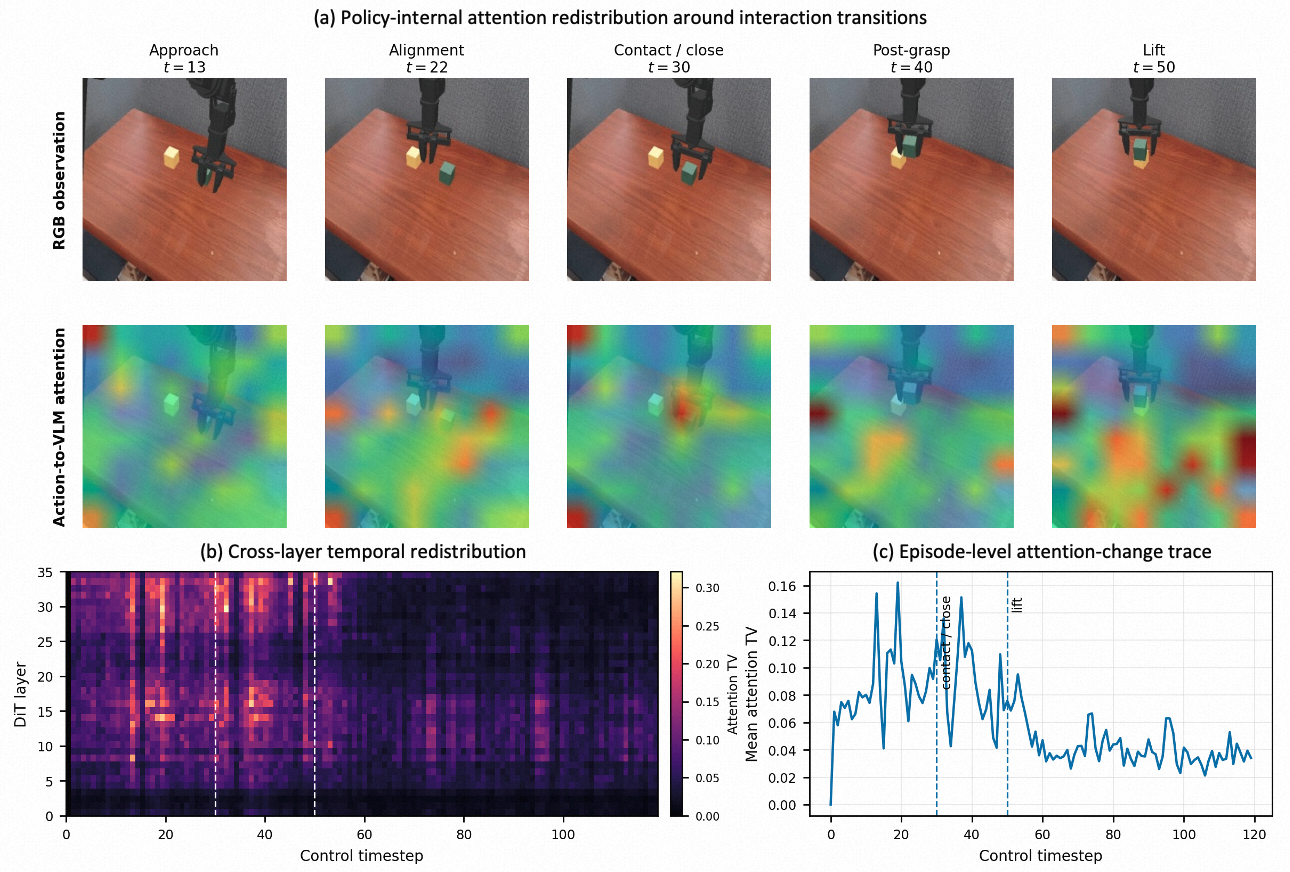}
    \caption{Policy-internal cross-layer attention changes around interaction
    transitions. (a) RGB observations (upper row) and representative
    action-query-to-VLM image-patch attention maps (lower row) from a Stack
    Blocks rollout at a fixed checkpoint; all attention maps share one
    episode-level scale. (b) Temporal redistribution of the action-to-VLM
    attention distribution across all 36 DiT layers. (c) The corresponding
    layer-averaged attention-change trace. Pronounced redistribution occurs
    around contact/closure and the subsequent post-grasp transition.}
    \label{fig:attention_event}
\end{figure}

\paragraph{Observation.}
\textbf{In the illustrated rollout, interaction transitions coincide with
pronounced cross-layer attention redistribution.}
In Figure~\ref{fig:attention_event}, panel (a) pairs observations from a
representative Stack Blocks rollout with the corresponding action-to-VLM
attention maps, while panels (b) and (c) show the layer-wise temporal
redistribution and its layer average, respectively. Attention varies throughout
the rollout, but marked
cross-layer redistributions occur around the labeled interaction transitions:
gripper--object contact and grasp closure coincide with pronounced changes
across multiple DiT layers, followed by another reorganization after grasping.
The event score measures redistribution of the full attention distribution
rather than the semantic location of its largest spatial response. This
single-rollout visualization is illustrative rather than a quantitative
event-detection benchmark; it does not establish detector precision or recall.
We therefore use attention change only as a label-free, policy-internal cue for
\emph{when to consider} a shadow update. Whether such a proposal is beneficial
is a separate question, tested by the future-validation ablation in
Sec.~\ref{sec:validation_ablation}.

\subsubsection{Visual-Representation Distribution Shift Across Evaluation Suites}
\label{sec:visual_shift_analysis}
\paragraph{Procedure.}
We use this analysis to compare the magnitudes of visual-representation shift
from the robot-data training reference for the WidowX and Google Robot
evaluation suites.
We first construct a reference set of 1,000 frames sampled equally from RT-1
~\cite{rt1} and Bridge~\cite{bridgedata_v2}; each evaluation suite
contributes another 1,000 frames. Every frame is passed through a fixed
checkpoint of the non-adapted QwenPi policy, and the last-layer Qwen3-VL image
tokens are mean-pooled into a 2,560-dimensional feature vector.

Next, we fit all preprocessing transforms on the reference set only and compare
two feature spaces: the raw standardized features (2,560 dimensions), and a
PCA space whose components retain 95\% of the reference variance (681
dimensions, denoted PCA-95). For each test suite, we compute the biased
RBF-MMD between its feature distribution and the reference distribution. MMD
is a kernel two-sample distance: a smaller value means that the test features
are closer to the training reference. The RBF bandwidth is selected from the
median pairwise distance within the reference set and kept fixed for both test
suites.

Finally, we quantify uncertainty by resampling complete episodes 500 times and
recomputing each MMD value, rather than resampling individual frames. The
brackets in Table~\ref{tab:representation_shift} are the resulting 95\% bootstrap
confidence intervals. Repeating the comparison in both feature spaces checks
that the ordering is not caused by dimensionality reduction alone.

\begin{table}[H]
\centering
\caption{Visual-representation distribution shift from the RT-1/Bridge training
mixture.}
\label{tab:representation_shift}
\small
\setlength{\tabcolsep}{3pt}
\begin{tabular}{lcc}
\toprule
Test suite & PCA-95 (681-D) & Raw standardized (2560-D) \\
\midrule
WidowX & 0.330 [0.326, 0.338] & 0.348 [0.343, 0.358] \\
Google Robot & 0.262 [0.259, 0.269] & 0.262 [0.260, 0.271] \\
\bottomrule
\end{tabular}
\end{table}

\paragraph{Observation.}
\textbf{WidowX is farther than Google Robot from the RT-1/Bridge visual-
representation reference in both evaluated feature spaces.}
Under this common feature extraction and kernel specification, WidowX is
farther from the RT-1/Bridge reference than Google Robot in both spaces:
$0.330$ versus $0.262$ after PCA-95 and $0.348$ versus $0.262$ in the raw
standardized space. We use this ordering only as distributional context. MMD
does not measure action semantics, embodiment dynamics, or closed-loop
difficulty, and it is not used to predict test-time training gains.

\FloatBarrier

\begingroup
\raggedbottom
\setlength{\intextsep}{12pt plus 2pt minus 2pt}

\subsection{Experimental Evaluation}
\label{sec:experiments}

\subsubsection{Experimental Setup}
\label{sec:experimental_setup}

\paragraph{Models and checkpoints.}
All variants use the same QwenPi backbone and robot-data training recipe
introduced in Sec.~\ref{sec:method}. We jointly train on Bridge
~\cite{bridgedata_v2} and RT-1~\cite{rt1} with equal sampling
probability, and compare state grounding or WPI with either
a single latent prompt or the context-routed Mixture of Latent Prompts
(MoLP). Unless a diagnostic explicitly uses a fixed checkpoint, results are
averaged over the consecutive 50K, 60K, 70K, and 80K checkpoints; evaluation
success is never used for checkpoint selection.

\paragraph{Benchmarks and closed-loop evaluation.}
We evaluate on the WidowX and Google Robot suites of SimplerEnv
\cite{simpler_env}. WidowX comprises Stack Green Cube on Yellow Cube, Put
Carrot on Plate, Put Spoon on Table Cloth, and Put Eggplant in Basket. Google
Robot comprises Pick Coke Can, Move Near, Open/Close Drawer, and Put in Drawer
under visual matching (VM) and variant aggregation (VA). At each control step,
overlapping action chunks are combined using the benchmark's adaptive temporal
ensemble, and only the resulting first action is executed before the next
observation. Compared variants enumerate the same explicit initial-condition
specifications, but we do not enforce a common-random-number rollout protocol
across methods. WidowX Overall is the unweighted mean over four tasks; Google
Overall is recomputed directly from the reported task-family--setting scores.

\paragraph{Adaptation protocols and notation.}
\emph{No TTT} leaves the trained prompt unchanged. \emph{Offline TTT} adapts
the prompt on a previously collected unlabeled suite buffer and then evaluates
the fixed adapted prompt. \emph{Online TTT} immediately deploys each prompt
update whenever a valid self-supervised pair becomes available. \emph{AGV-TTT}
proposes updates only at attention events and deploys a candidate only after
future validation. In the comparison tables, SG-LP denotes the state-grounded
single latent prompt used by QwenPi, while FP-LP denotes the future-predictive
single latent prompt used by QwenWPI; FP-LP contains no state expert. Unless
stated otherwise, TTT in the public-comparison tables follows the Offline TTT
protocol of TTT-VLA~\cite{ttt-vla}. Optimization settings are detailed in
Appendix~\ref{app:training_details}, and the gating and validation settings in
Appendix~\ref{app:agv_details}.

\subsubsection{Comparison with Prior Policies}
\label{sec:sota_comparison}

\begin{table}[H]
    \centering
    \caption{Comparison with prior policies on the SimplerEnv WidowX benchmark.}
    \label{tab:widowx_sota}
    \small
    \renewcommand{\arraystretch}{1.05}
    \setlength{\tabcolsep}{4.5pt}
    \begin{tabular*}{\textwidth}{@{\extracolsep{\fill}}lccccc@{}}
        \toprule
        \textbf{Method}
        & \textbf{Carrot}
        & \textbf{Eggplant}
        & \textbf{Spoon}
        & \textbf{Cube}
        & \textbf{Overall} \\
        \midrule
        RT-1-X~\cite{openx} & 4.2 & 0.0 & 0.0 & 0.0 & 1.1 \\
        OpenVLA~\cite{openvla} & 0.0 & 4.1 & 0.0 & 0.0 & 1.0 \\
        SpatialVLA~\cite{spatialvla} & 20.8 & 70.8 & 20.8 & 25.0 & 34.4 \\
        Magma~\cite{magma} & 29.2 & 91.7 & 37.5 & 20.8 & 44.8 \\
        Octo-Base~\cite{octo} & 8.3 & 43.1 & 12.5 & 0.0 & 16.0 \\
        Octo-Small~\cite{octo} & 9.7 & 56.9 & 47.2 & 4.2 & 29.5 \\
        RoboVLMs~\cite{robovlms} & 20.8 & 79.2 & 45.8 & 4.2 & 37.5 \\
        InstructVLA~\cite{instructvla} & 40.3 & 94.4 & 43.1 & 9.7 & 46.9 \\
        $\pi_0$~\cite{pi0} & 36.1 & 81.9 & 45.8 & 26.4 & 47.6 \\
        CogACT~\cite{cogact} & 37.5 & 91.7 & 58.3 & 20.8 & 52.1 \\
        ThinkAct~\cite{thinkact} & 37.5 & 70.8 & 58.3 & 8.7 & 43.8 \\
        TTT-VLA ($\pi_{0.5}$ + SG-LP + TTT)~\cite{ttt-vla}
        & \textbf{74.5} & 76.0 & 71.0 & \textbf{48.0} & 67.4 \\
        \midrule
        QwenPi + SG-LP + TTT
        & 58.0 & 95.6 & 83.9 & 18.2 & 63.9 \\
        QwenPi + MoLP + AGV-TTT
        & 64.6 & \textbf{97.1} & \textbf{84.9} & 28.4 & 68.8 \\
        \addlinespace[3pt]
        QwenWPI + FP-LP + TTT
        & 63.1 & 90.2 & 73.9 & 44.9 & 68.0 \\
        QwenWPI + MoLP + AGV-TTT
        & 68.8 & 94.5 & 77.3 & 44.3 & \textbf{71.2} \\
        \bottomrule
    \end{tabular*}
\end{table}

\paragraph{WidowX.}
Table~\ref{tab:widowx_sota} reports task-level success and the unweighted
four-task average. Values for prior policies follow the public comparison in
TTT-VLA~\cite{ttt-vla}; our rows average the four consecutive checkpoints.
Within the QwenPi family, replacing SG-LP + TTT with MoLP + AGV-TTT raises
Overall from $63.9\%$ to $68.8\%$; within QwenWPI, the corresponding end-to-end
comparison improves from $68.0\%$ to $71.2\%$. The latter also exceeds the
published TTT-VLA result of $67.4\%$ by $3.8$ points. These comparisons show the
performance of the complete VANE configuration against the corresponding TTT
baselines. Because each pair changes both prompt structure and update protocol,
and the cross-system comparison additionally changes backbone and training data,
we defer component attribution to Sec.~\ref{sec:controlled_evaluation}.

\begin{table}[H]
    \centering
    \caption{Comparison with prior policies on three common Google Robot task families.}
    \label{tab:google_sota}
    \small
    \renewcommand{\arraystretch}{1.05}
    \setlength{\tabcolsep}{2.6pt}
    \begin{tabular*}{\textwidth}{@{\extracolsep{\fill}}lccccccc@{}}
        \toprule
        \multirow{2}{*}{\textbf{Method}}
        & \multicolumn{2}{c}{\textbf{Pick Coke Can}}
        & \multicolumn{2}{c}{\textbf{Move Near}}
        & \multicolumn{2}{c}{\textbf{Open/Close Drawer}}
        & \multirow{2}{*}{\textbf{Overall}} \\
        \cmidrule(lr){2-3}\cmidrule(lr){4-5}\cmidrule(lr){6-7}
        & \textbf{VM} & \textbf{VA}
        & \textbf{VM} & \textbf{VA}
        & \textbf{VM} & \textbf{VA}
        & \\
        \midrule
        TraceVLA~\cite{tracevla}
        & 28.0 & 60.0 & 53.7 & 56.4 & 57.0 & 31.0 & 47.7 \\
        RT-1-X~\cite{openx}
        & 56.7 & 49.0 & 31.7 & 32.3 & 59.7 & 29.4 & 43.1 \\
        Octo-Base~\cite{octo}
        & 17.0 & 0.0 & 4.2 & 3.1 & 22.7 & 1.1 & 8.0 \\
        OpenVLA~\cite{openvla}
        & 16.3 & 54.5 & 46.2 & 47.7 & 35.6 & 17.7 & 36.3 \\
        RoboVLMs (zero-shot)~\cite{robovlms}
        & 72.7 & 68.3 & 66.3 & 56.0 & 26.8 & 8.5 & 49.8 \\
        RoboVLMs (fine-tuned)~\cite{robovlms}
        & 77.3 & 75.6 & 61.7 & 60.0 & 43.5 & 10.6 & 54.8 \\
        Emma-X~\cite{emmax}
        & 2.3 & 5.3 & 3.3 & 7.3 & 18.3 & 20.5 & 9.5 \\
        $\pi_0$ (fine-tuned)~\cite{pi0}
        & 72.7 & 75.2 & 65.3 & 63.7 & 38.3 & 25.6 & 56.8 \\
        $\pi_0$-FAST (fine-tuned)~\cite{pi0fast}
        & 75.3 & 77.6 & 67.5 & 68.2 & 42.9 & 31.3 & 60.5 \\
        GR00T (fine-tuned)~\cite{groot}
        & 69.3 & 46.7 & 68.7 & 62.9 & 35.8 & 17.5 & 50.1 \\
        TTT-VLA ($\pi_{0.5}$ + SG-LP + TTT)~\cite{ttt-vla}
        & 85.0 & 79.3 & 71.7 & 55.2 & 60.6 & 45.8 & 66.3 \\
        \midrule
        QwenPi + SG-LP + TTT
        & 87.6 & 78.7 & 79.3 & 66.9 & \textbf{68.2} & 49.3 & 71.7 \\
        QwenPi + MoLP + AGV-TTT
        & \textbf{94.8} & \textbf{85.8} & \textbf{87.5} & \textbf{71.9}
        & 65.3 & 55.1 & \textbf{76.7} \\
        \addlinespace[3pt]
        QwenWPI + FP-LP + TTT
        & 84.3 & 82.4 & 81.6 & 68.8 & 64.8 & \textbf{59.8} & 73.6 \\
        QwenWPI + MoLP + AGV-TTT
        & 89.1 & 83.7 & 83.3 & 67.5 & 55.2 & 55.4 & 72.4 \\
        \bottomrule
    \end{tabular*}
\end{table}

\paragraph{Google Robot.}
The public comparison in TTT-VLA reports Pick Coke Can, Move Near, and
Open/Close Drawer but not Put in Drawer. Table~\ref{tab:google_sota} therefore
uses the same three-task scope and averages the six VM/VA entries for Overall.
Within the QwenPi family, the complete VANE configuration raises Overall from
$71.7\%$ for SG-LP + TTT to $76.7\%$ for MoLP + AGV-TTT. This is an end-to-end
comparison, not an estimate of either component's isolated gain. Within QwenWPI,
the corresponding result decreases from $73.6\%$ to $72.4\%$, indicating that
future-predictive prompting is not uniformly superior across this restricted
task aggregation. The strongest VANE variant, QwenPi + MoLP + AGV-TTT, exceeds
the published TTT-VLA result of $66.3\%$ by $10.4$ points, but this cross-system
difference does not isolate backbone, training-data, or adaptation effects. We
therefore treat the Google comparison as a boundary and scope test rather than
evidence of uniform superiority across embodiments.

\begin{table}[H]
    \centering
    \caption{Complete four-task breakdown of our variants on Google Robot.}
    \label{tab:google_full}
    \footnotesize
    \renewcommand{\arraystretch}{1.05}
    \setlength{\tabcolsep}{1.8pt}
    \begin{tabular*}{\textwidth}{@{\extracolsep{\fill}}lccccccccc@{}}
        \toprule
        \multirow{2}{*}{\textbf{Method}}
        & \multicolumn{2}{c}{\textbf{Pick Coke Can}}
        & \multicolumn{2}{c}{\textbf{Move Near}}
        & \multicolumn{2}{c}{\textbf{Open/Close Drawer}}
        & \multicolumn{2}{c}{\textbf{Put in Drawer}}
        & \multirow{2}{*}{\textbf{Overall}} \\
        \cmidrule(lr){2-3}\cmidrule(lr){4-5}\cmidrule(lr){6-7}\cmidrule(lr){8-9}
        & \textbf{VM} & \textbf{VA}
        & \textbf{VM} & \textbf{VA}
        & \textbf{VM} & \textbf{VA}
        & \textbf{VM} & \textbf{VA}
        & \\
        \midrule
        QwenPi + SG-LP + TTT
        & 87.6 & 78.7 & 79.3 & 66.9 & \textbf{68.2} & 49.3
        & \textbf{72.0} & 63.4 & 70.7 \\
        QwenPi + MoLP + AGV-TTT
        & \textbf{94.8} & \textbf{85.8} & \textbf{87.5} & \textbf{71.9}
        & 65.3 & 55.1 & 60.2 & 52.9 & \textbf{71.7} \\
        \addlinespace[3pt]
        QwenWPI + FP-LP + TTT
        & 84.3 & 82.4 & 81.6 & 68.8 & 64.8 & \textbf{59.8}
        & 19.0 & 56.0 & 64.6 \\
        QwenWPI + MoLP + AGV-TTT
        & 89.1 & 83.7 & 83.3 & 67.5 & 55.2 & 55.4
        & 58.8 & \textbf{70.4} & 70.4 \\
        \bottomrule
    \end{tabular*}
\end{table}

\paragraph{Complete four-task breakdown.}
Table~\ref{tab:google_full} adds Put in Drawer and averages all eight
VM/VA task--setting entries. Within QwenPi, MoLP + AGV-TTT raises Overall from
$70.7\%$ to $71.7\%$ relative to SG-LP + TTT. Within QwenWPI, the complete VANE
configuration rises markedly from $64.6\%$ to $70.4\%$ after Put in Drawer is
included. The largest task-family difference between the two QwenWPI rows occurs
on Put in Drawer, where VM/VA increase from $19.0\%/56.0\%$ to
$58.8\%/70.4\%$. Together with the restricted three-task result, this shows that
the end-to-end benefit of MoLP and AGV-TTT depends on task composition. Because
WPI is held fixed within that comparison, the result does not attribute the
composition effect to the predictive proxy. We therefore treat Google Robot as
a scope test rather than evidence that one proxy or adaptation protocol
uniformly dominates every task family.

\subsubsection{Controlled Evaluation of VANE Components}
\label{sec:controlled_evaluation}

\begin{table}[H]
\centering
\caption{Controlled comparison on WidowX over four checkpoints.}
\label{tab:widowx_main}
\small
\renewcommand{\arraystretch}{1.08}
\setlength{\tabcolsep}{6pt}
\begin{tabular}{llcccc}
\toprule
Proxy & Prompt & No TTT & Offline & Online & AGV \\
\midrule
State & Single & 64.3 & 63.9 & 64.7 & 65.9 \\
State & MoLP   & 67.5 & 68.1 & 67.8 & 68.8 \\
WPI   & Single & 67.5 & 68.0 & 67.5 & 69.0 \\
WPI   & MoLP   & 70.7 & 69.4 & 70.3 & \textbf{71.2} \\
\bottomrule
\end{tabular}
\end{table}

\paragraph{WidowX.}
Table~\ref{tab:widowx_main} provides the factorial comparison used for
component attribution. The strongest value is $71.2\%$ for WPI--MoLP with
AGV-TTT, but the important evidence is the matched comparison along one factor
at a time, summarized below.

\begin{table}[H]
\centering
\caption{Controlled comparison on the four-task Google Robot suite.}
\label{tab:google_main}
\small
\renewcommand{\arraystretch}{1.08}
\setlength{\tabcolsep}{6pt}
\begin{tabular}{llcccc}
\toprule
Proxy & Prompt & No TTT & Offline & Online & AGV \\
\midrule
State & Single & 73.6 & 70.7 & 74.0 & 74.0 \\
State & MoLP   & 72.3 & 71.7 & 71.6 & 71.7 \\
WPI   & Single & 64.2 & 64.6 & 64.7 & 65.3 \\
WPI   & MoLP   & 70.1 & 70.2 & 70.5 & 70.4 \\
\bottomrule
\end{tabular}
\end{table}

\paragraph{Google Robot.}
Table~\ref{tab:google_main} provides the same factorial comparison on the
complete Google suite, where the component ranking changes and the benefits are
less uniform. In particular, the strongest configuration remains the
state-grounded single-prompt policy, while both WPI rows receive positive but
smaller AGV gains over No TTT. Thus, the controlled component benefits are more
consistent on WidowX than on Google Robot. These outcomes should be interpreted
alongside, rather than explained by, the descriptive representation-shift
analysis in Sec.~\ref{sec:visual_shift_analysis}.

\paragraph{Prompt structure.}
On WidowX, MoLP improves every matched single-prompt configuration when the
proxy and adaptation protocol are held fixed. For state grounding, the gains
range from $+2.9$ to $+4.2$ points across No TTT, Offline, Online, and AGV; for
WPI, the corresponding gains range from $+1.4$ to $+3.2$ points. The positive
No-TTT comparisons show that the routed prompt structure already changes the
trained policy before deployment-time optimization, while the remaining columns
show that the advantage persists under the evaluated TTT protocols. Google
Robot is less uniform: MoLP consistently improves the WPI variants but not the
state-grounded variants. Together with the diagnostic task-sharing analysis in
Sec.~\ref{sec:task_interference_analysis}, these comparisons separate the
motivation for routed prompts from their measured effect.

\paragraph{Predictive proxy.}
On WidowX, WPI exceeds state grounding in all eight matched prompt--protocol
comparisons. The No-TTT gap is already $+3.2$ points for both Single and MoLP,
showing that a substantial part of the absolute WPI advantage is present before
deployment-time optimization. Under Offline, Online, and AGV, the corresponding
absolute gaps remain positive at $1.3$--$4.1$ points. These comparisons therefore
support WPI as a stronger proxy-conditioned policy interface on WidowX, but they
do not by themselves establish that WPI increases the incremental gain from
TTT: improvements relative to each model's own No-TTT baseline remain
protocol-dependent. On Google Robot, WPI does not exceed state grounding in
any of the eight matched comparisons, further bounding the proxy claim to the
evaluated task and embodiment distribution.

\paragraph{Deployment protocol.}
On WidowX, AGV-TTT is the best evaluated protocol for every proxy--prompt row
and the only one that also improves all four rows over their corresponding
No-TTT baselines. Relative to No TTT, the gains are $+1.6$, $+1.3$, $+1.5$,
and $+0.5$ points for State--Single, State--MoLP, WPI--Single, and WPI--MoLP,
respectively; relative to dense Online TTT, AGV improves the same four rows by
$+1.2$, $+1.0$, $+1.5$, and $+0.9$ points. On Google Robot, the ordering is not
uniform: AGV improves both WPI rows over No TTT, slightly improves State--Single,
and degrades State--MoLP. This motivates isolating the validation step from
event-triggered sparsity in Sec.~\ref{sec:validation_ablation}.

\subsubsection{Is Future Validation Necessary?}
\label{sec:validation_ablation}
\begin{table}[H]
\centering
\caption{Ablation of event triggering and future validation on single-prompt QwenWPI.}
\label{tab:accept_all}
\small
\setlength{\tabcolsep}{5.0pt}
\begin{tabular}{lccc}
\toprule
Online protocol & Event gate & Future validation & Success \\
\midrule
Online TTT          & --           & --           & 67.5 \\
Event + Accept-All  & $\checkmark$ & --           & 67.3 \\
AGV-TTT             & $\checkmark$ & $\checkmark$ & \textbf{69.0} \\
\bottomrule
\end{tabular}
\end{table}

\paragraph{Result.}
To separate future validation from event triggering, \emph{Accept-All} retains
the same attention detector and shadow proposal as AGV-TTT but commits every
completed candidate without evaluating it on subsequent observations. As shown
in Table~\ref{tab:accept_all}, Event + Accept-All reaches $67.3\%$, slightly
below Online TTT at $67.5\%$. Future validation raises success to $69.0\%$, a
$1.7$-point improvement over Accept-All. Event-triggered sparsity alone
therefore does not explain the gain. The $1.7$-point gap isolates a measurable
benefit of future validation within the same event-gated proposal pipeline,
consistent with validation filtering unfavorable candidates before deployment.

\subsubsection{Optimization Efficiency}
\label{sec:optimization_efficiency}
\begin{table}[H]
\centering
\caption{Per-checkpoint optimization and validation cost for single-prompt QwenWPI, averaged over four checkpoints.}
\label{tab:update_efficiency}
\small
\setlength{\tabcolsep}{4.0pt}
\begin{tabular}{lrrr}
\toprule
Protocol & Backward & Validation forward & Success \\
\midrule
Online TTT & 45,824 & -- & 67.5 \\
AGV-TTT & 612 & 4,831 & \textbf{69.0} \\
\bottomrule
\end{tabular}
\end{table}

\paragraph{Result.}
Online TTT performs a backward update whenever a valid WPI pair becomes
available, whereas AGV-TTT proposes a shadow update only at an attention event.
For the same single-prompt QwenWPI policy over four checkpoints,
Table~\ref{tab:update_efficiency} shows that AGV-TTT performs an average of
$612$ candidate backward passes per checkpoint evaluation,
compared with $45{,}824$ for Online TTT. It triggers backward optimization on only $1.3\%$ of eligible
frames and reduces backward calls by $98.7\%$ (approximately $75\times$), while
improving mean success from $67.5\%$ to $69.0\%$.

AGV-TTT additionally performs $4{,}831$ forward-only proxy evaluations for
future validation. Only $19.4\%$ of proposals are accepted, so most shadow
proposals never enter the deployed policy. This selectivity is functionally important:
Event + Accept-All obtains $67.3\%$, whereas validated AGV-TTT reaches
$69.0\%$. The gain therefore comes from event-triggered proposals together
with subsequent validation, not from reducing update frequency alone. We
report optimization counts rather than converting them into an unsupported
wall-clock speedup.

\FloatBarrier
\endgroup

\section{CONCLUSION}
We presented VANE, a test-time training framework for VLAs that combines a
context-routed MoLP, future visual representation prediction,
and attention-gated, validation-driven updates. Attention redistribution
identifies informative interaction transitions for proposing shadow updates,
while subsequent observations determine whether each candidate is committed or
rolled back.

On SimplerEnv WidowX, the learned WPI--MoLP interface improves the non-adapted
policy from $64.3\%$ to $70.7\%$, and AGV-TTT further raises success to
$71.2\%$. AGV-TTT is the only evaluated protocol that improves
checkpoint-averaged performance across all four proxy--prompt configurations.
In the single-prompt QwenWPI efficiency study, it reduces backward updates by
$98.7\%$ relative to Online TTT. Google Robot results further show that
adaptation gains remain task- and embodiment-dependent, alongside measurable
differences in visual-representation shift. These findings support a context-routed, temporally predictive adaptation
interface on WidowX and an explicit proposal--validation mechanism for online
updates, while the Google results bound the universality of these gains.

\bibliographystyle{IEEEtran}
\bibliography{references}

@inproceedings{ttt,
  title={Test-Time Training with Self-Supervision for Generalization under Distribution Shifts},
  author={Sun, Yu and Wang, Xiaolong and Liu, Zhuang and Miller, John and Efros, Alexei A. and Hardt, Moritz},
  booktitle={Proceedings of the 37th International Conference on Machine Learning},
  pages={9229--9248},
  year={2020}
}

@article{tent,
  title={Tent: Fully Test-Time Adaptation by Entropy Minimization},
  author={Wang, Dequan and Shelhamer, Evan and Liu, Shaoteng and Olshausen, Bruno and Darrell, Trevor},
  journal={arXiv preprint arXiv:2006.10726},
  year={2020}
}

@article{cotta,
  title={Continual Test-Time Domain Adaptation},
  author={Wang, Qin and Fink, Olga and Van Gool, Luc and Dai, Dengxin},
  journal={arXiv preprint arXiv:2203.13591},
  year={2022}
}

@article{eata,
  title={Efficient Test-Time Model Adaptation without Forgetting},
  author={Niu, Shuaicheng and Wu, Jiaxiang and Zhang, Yifan and Chen, Yaofo and Zheng, Shijian and Zhao, Peilin and Tan, Mingkui},
  journal={arXiv preprint arXiv:2204.02610},
  year={2022}
}

@inproceedings{ttt-llm,
  title={Test-Time Training on Nearest Neighbors for Large Language Models},
  author={Hardt, Moritz and Sun, Yu},
  booktitle={International Conference on Learning Representations},
  year={2024}
}

@inproceedings{tpt,
  title={Test-Time Prompt Tuning for Zero-Shot Generalization in Vision-Language Models},
  author={Shu, Manli and Nie, Weili and Huang, De-An and Yu, Zhiding and Goldstein, Tom and Anandkumar, Anima and Xiao, Chaowei},
  booktitle={Advances in Neural Information Processing Systems},
  year={2022}
}

@article{rt1,
  title={{RT-1}: Robotics Transformer for Real-World Control at Scale},
  author={Brohan, Anthony and others},
  journal={arXiv preprint arXiv:2212.06817},
  year={2022}
}

@article{rt2,
  title={{RT-2}: Vision-Language-Action Models Transfer Web Knowledge to Robotic Control},
  author={Brohan, Anthony and others},
  journal={arXiv preprint arXiv:2307.15818},
  year={2023}
}

@article{openx,
  title={{Open X-Embodiment}: Robotic Learning Datasets and {RT-X} Models},
  author={{Open X-Embodiment Collaboration} and O'Neill, Abby and others},
  journal={arXiv preprint arXiv:2310.08864},
  year={2023}
}

@article{bridgedata_v2,
  title={{BridgeData V2}: A Dataset for Robot Learning at Scale},
  author={Walke, Homer and Black, Kevin and Lee, Abraham and Kim, Moo Jin and Du, Max and Zheng, Chongyi and Zhao, Tony and Hansen-Estruch, Philippe and Vuong, Quan and He, Andre and Myers, Vivek and Fang, Kuan and Finn, Chelsea and Levine, Sergey},
  journal={arXiv preprint arXiv:2308.12952},
  year={2023}
}

@article{octo,
  title={Octo: An Open-Source Generalist Robot Policy},
  author={{Octo Model Team} and Ghosh, Dibya and others},
  journal={arXiv preprint arXiv:2405.12213},
  year={2024}
}

@article{openvla,
  title={{OpenVLA}: An Open-Source Vision-Language-Action Model},
  author={Kim, Moo Jin and others},
  journal={arXiv preprint arXiv:2406.09246},
  year={2024}
}

@article{pi0,
  title={$\pi_0$: A Vision-Language-Action Flow Model for General Robot Control},
  author={Black, Kevin and others},
  journal={arXiv preprint arXiv:2410.24164},
  year={2024}
}

@article{starVLA-alpha,
  title={{StarVLA}-$\alpha$: Reducing Complexity in Vision-Language-Action Systems},
  author={Ye, Jinhui and Gao, Ning and Yang, Senqiao and Zheng, Jinliang and Wang, Zixuan and Chen, Yuxin and Chen, Pengguang and Chen, Yilun and Liu, Shu and Jia, Jiaya},
  journal={arXiv preprint arXiv:2604.11757},
  year={2026}
}

@article{ttt-vla,
  title={{TTT-VLA}: Test-Time Latent Prompt Optimization for Vision-Language-Action Models},
  author={Zhang, Wenbo and Li, Jianxiong and Yang, Shuai and Chen, Sijin and Liu, Jiajun and Liu, Lingqiao and Ma, Xiao},
  journal={arXiv preprint arXiv:2606.03127},
  year={2026}
}

@article{tt-vla,
  title={On-the-Fly {VLA} Adaptation via Test-Time Reinforcement Learning},
  author={Liu, Changyu and Liu, Yiyang and Wang, Taowen and Zhuang, Qiao and Liang, James Chenhao and Yang, Wenhao and Xu, Renjing and Wang, Qifan and Liu, Dongfang and Han, Cheng},
  journal={arXiv preprint arXiv:2601.06748},
  year={2026}
}

@article{t3vf,
  title={Test-Time Training for Visual Foresight Vision-Language-Action Models},
  author={Park, Sangwu and Kim, Wonjoong and In, Yeonjun and Kim, Sein and Kang, Hongseok and Park, Chanyoung},
  journal={arXiv preprint arXiv:2605.08215},
  year={2026}
}

@article{vjepa2,
  title={{V-JEPA 2}: Self-Supervised Video Models Enable Understanding, Prediction and Planning},
  author={Assran, Mido and others},
  journal={arXiv preprint arXiv:2506.09985},
  year={2025}
}

@article{futurevla,
  title={{FutureVLA}: Joint Visuomotor Prediction for Vision-Language-Action Model},
  author={Xu, Xiaoxu and Li, Hao and Ye, Jinhui and Chen, Yilun and Zeng, Jia and Chen, Xinyi and Xu, Linning and Lin, Dahua and Li, Weixin and Pang, Jiangmiao},
  journal={arXiv preprint arXiv:2603.10712},
  year={2026}
}

@article{lingbot-va,
  title={Causal World Modeling for Robot Control},
  author={Li, Lin and Zhang, Qihang and Luo, Yiming and Yang, Shuai and Wang, Ruilin and Han, Fei and Yu, Mingrui and Gao, Zelin and Xue, Nan and Zhu, Xing and Shen, Yujun and Xu, Yinghao},
  journal={arXiv preprint arXiv:2601.21998},
  year={2026}
}

@article{simpler_env,
  title={Evaluating Real-World Robot Manipulation Policies in Simulation},
  author={Li, Xuanlin and Hsu, Kyle and Gu, Jiayuan and Pertsch, Karl and Mees, Oier and Walke, Homer Rich and Fu, Chuyuan and Lunawat, Ishikaa and Sieh, Isabel and Kirmani, Sean and Levine, Sergey and Wu, Jiajun and Finn, Chelsea and Su, Hao and Vuong, Quan and Xiao, Ted},
  journal={arXiv preprint arXiv:2405.05941},
  year={2024}
}

@article{spatialvla,
  title   = {{SpatialVLA}: Exploring Spatial Representations for
             Vision-Language-Action Model},
  author  = {Qu, Delin and Song, Haoming and Chen, Qizhi and Yao, Yuanqi
             and Ye, Xinyi and Ding, Yan and Wang, Zhigang and Gu, JiaYuan
             and Zhao, Bin and Wang, Dong and Li, Xuelong},
  journal = {arXiv preprint arXiv:2501.15830},
  year    = {2025}
}

@inproceedings{magma,
  title     = {Magma: A Foundation Model for Multimodal {AI} Agents},
  author    = {Yang, Jianwei and Tan, Reuben and Wu, Qianhui and
               Zheng, Ruijie and Peng, Baolin and Liang, Yongyuan and
               Gu, Yu and Cai, Mu and Ye, Seonghyeon and Jang, Joel and
               Deng, Yuquan and Gao, Jianfeng},
  booktitle = {Proceedings of the IEEE/CVF Conference on Computer Vision
               and Pattern Recognition},
  pages     = {14203--14214},
  year      = {2025}
}

@article{robovlms,
  title   = {Towards Generalist Robot Policies: What Matters in Building
             Vision-Language-Action Models},
  author  = {Li, Xinghang and Li, Peiyan and Liu, Minghuan and Wang, Dong
             and Liu, Jirong and Kang, Bingyi and Ma, Xiao and Kong, Tao
             and Zhang, Hanbo and Liu, Huaping},
  journal = {arXiv preprint arXiv:2412.14058},
  year    = {2024}
}

@article{instructvla,
  title   = {{InstructVLA}: Vision-Language-Action Instruction Tuning
             from Understanding to Manipulation},
  author  = {Yang, Shuai and Li, Hao and Wang, Bin and Chen, Yilun and
             Tian, Yang and Wang, Tai and Wang, Hanqing and Zhao, Feng
             and Liao, Yiyi and Pang, Jiangmiao},
  journal = {arXiv preprint arXiv:2507.17520},
  year    = {2025}
}

@article{cogact,
  title   = {{CogACT}: A Foundational Vision-Language-Action Model for
             Synergizing Cognition and Action in Robotic Manipulation},
  author  = {Li, Qixiu and Liang, Yaobo and Wang, Zeyu and Luo, Lin and
             Chen, Xi and Liao, Mozheng and Wei, Fangyun and Deng, Yu and
             Xu, Sicheng and Zhang, Yizhong and Wang, Xiaofan and Liu, Bei
             and Fu, Jianlong and Bao, Jianmin and Chen, Dong and Shi,
             Yuanchun and Yang, Jiaolong and Guo, Baining},
  journal = {arXiv preprint arXiv:2411.19650},
  year    = {2024}
}

@inproceedings{thinkact,
  title     = {{ThinkAct}: Vision-Language-Action Reasoning via Reinforced
               Visual Latent Planning},
  author    = {Huang, Chi-Pin and Wu, Yueh-Hua and Chen, Min-Hung and
               Wang, Yu-Chiang Frank and Yang, Fu-En},
  booktitle = {Advances in Neural Information Processing Systems},
  year      = {2025}
}

@inproceedings{tracevla,
  title={{TraceVLA}: Visual Trace Prompting Enhances Spatial-Temporal Awareness for Generalist Robotic Policies},
  author={Zheng, Ruijie and Liang, Yongyuan and Huang, Shuaiyi and Gao, Jianfeng and Daum\'e III, Hal and Kolobov, Andrey and Huang, Furong and Yang, Jianwei},
  booktitle={International Conference on Learning Representations},
  year={2025}
}

@article{emmax,
  title={{Emma-X}: An Embodied Multimodal Action Model with Grounded Chain of Thought and Look-ahead Spatial Reasoning},
  author={Sun, Qi and Hong, Pengfei and Pala, Tej Deep and Toh, Vernon and Tan, U-Xuan and Ghosal, Deepanway and Poria, Soujanya},
  journal={arXiv preprint arXiv:2412.11974},
  year={2024}
}

@inproceedings{pi0fast,
  title={{FAST}: Efficient Action Tokenization for Vision-Language-Action Models},
  author={Pertsch, Karl and Stachowicz, Kyle and Ichter, Brian and Driess, Danny and Nair, Suraj and Vuong, Quan and Mees, Oier and Finn, Chelsea and Levine, Sergey},
  booktitle={Proceedings of Robotics: Science and Systems},
  year={2025}
}

@article{groot,
  title={{GR00T N1}: An Open Foundation Model for Generalist Humanoid Robots},
  author={{NVIDIA} and Bjorck, Johan and Castaneda, Fernando and Cherniadev, Nikita and Da, Xingye and Ding, Runyu and Fan, Linxi and Fang, Yu and Fox, Dieter and others},
  journal={arXiv preprint arXiv:2503.14734},
  year={2025}
}

\clearpage
\appendix
\section*{APPENDIX}
\makeatletter
\renewcommand{\@seccntformat}[1]{\csname the#1\endcsname.\ }
\makeatother

\section{Notation}
\label{app:notation}
The table summarizes symbols reused across the method and deployment procedure;
quantities local to a single derivation are defined where they are introduced.
\begin{center}
\scriptsize
\setlength{\tabcolsep}{3.5pt}
\renewcommand{\arraystretch}{1.10}
\begin{tabular}{lp{0.72\columnwidth}}
\toprule
Symbol & Definition \\
\midrule
$o_t,\ell$ & RGB observation and language instruction at control step $t$. \\
$\widehat A_t,a_t$ & Policy-predicted action chunk and executed action at control step $t$. \\
$\Theta$ & Policy parameters, frozen during deployment-time adaptation. \\
$\mathcal P,P_i,p_t$ & Latent-prompt bank, prompt component $i$, and effective routed prompt. \\
$c_t,q_t,\mathcal S_t$ & Router context, prompt-selection probabilities, and selected set. \\
$E,K$ & Number of banked latent prompts and selected prompts. \\
$f_\omega,Z_t,Z_{t+k}$ & Frozen visual encoder and its present and delayed future visual tokens. \\
$\mathcal L_{\mathrm{act}},\mathcal L_{\mathrm{wpi}},\mathcal L_{\mathrm{lb}}$ & Action, future-prediction, and router load-balancing losses. \\
$a^c_{t,u,l},\Delta_t^c$ & Channel-$c$ attention at step $t$, flow step $u$, and layer $l$, and its aggregated temporal change. \\
$\mathcal K_t,F_{t,n},e_{t,n}$ & Focused token set, normalized weight, and token-wise prediction error. \\
$\mathcal L_{\mathrm{focus}}$ & Attention-weighted prediction loss over focused tokens. \\
$\mathcal P^{\mathrm{old}},\mathcal P^{\mathrm{cand}}$ & Live and shadow candidate prompt banks. \\
$H_v,R_x$ & Validation horizon and relative improvement for loss $x$. \\
\bottomrule
\end{tabular}
\end{center}

\section{Training and Architectural Details}
\label{app:training_details}

\paragraph{Flow-matching action objective.}
For the clean action chunk $A_1\in\mathbb R^{16\times7}$, we sample
$A_0\sim\mathcal N(0,I)$ and $u\sim\operatorname{Beta}(1.5,1.0)$, then set
\begin{equation}
    \tau=\frac{\eta-u}{\eta},\quad \eta=0.999,\qquad
    A_\tau=(1-\tau)A_0+\tau A_1.
    \label{eq:app_flow_path}
\end{equation}
Let $H_a$ and $d_a$ denote the action-chunk horizon and per-step action
dimension. The action-branch output $\widehat v_\Theta$ predicts the constant
target velocity $A_1-A_0$:
\begin{equation}
    \mathcal L_{\mathrm{act}}
    =\mathbb E\left[
    \frac{1}{H_a d_a}
    \left\|\widehat v_\Theta(A_\tau,\tau)-(A_1-A_0)\right\|_F^2
    \right].
    \label{eq:app_action_loss}
\end{equation}
All action steps and dimensions are weighted equally. Repeated noise-time
samples provide a Monte Carlo estimate of the same objective.

\paragraph{Router regularization and initialization.}
For a batch of samples, let $f_i$ be the fraction routed to latent prompt $i$ and
$\bar q_i$ its mean router probability. We use
\begin{equation}
    \mathcal L_{\mathrm{lb}}=E\sum_{i=1}^{E}f_i\bar q_i,
    \quad
    \mathcal L_{\mathrm{train}}
    =\mathcal L_{\mathrm{act}}+0.1\mathcal L_{\mathrm{wpi}}
     +0.01\mathcal L_{\mathrm{lb}}.
    \label{eq:app_load_balance}
\end{equation}
The latent-prompt bank has shape $[8,16,1024]$; its eight prompt components are
independently initialized from $\mathcal N(0,0.02^2)$. The linear router uses the PyTorch default
initialization, and the one-stage model is not restored from an earlier prompt
or router checkpoint.

\paragraph{Latent-Action DiT visibility.}
The WPI query is ordered as
$[\text{prompt}\mid\text{present visual latent}\mid\text{action}]$,
with bidirectional interaction between present-latent and action tokens.
Following the latent-prompt visibility strategy of
TTT-VLA~\cite{ttt-vla}, the direct action-query-to-prompt-key connection is
masked during prompt learning and restored during policy inference and
prompt-only TTT. This inherited design prevents the action branch from relying
prematurely on the latent prompt during training, while allowing the adapted
prompt to modulate the frozen action branch at deployment. QwenWPI changes the
proxy interface by introducing predictive visual-latent tokens, but does not
alter this prompt-to-action visibility strategy. Prompt/visual and action
tokens use separate feed-forward branches while sharing attention. V-JEPA 2
remains frozen, the future offset is $k=8$, and the WPI loss averages ordinary
squared error over all visual tokens and channels.

\paragraph{Prompt-only optimization.}
At test time, the full $[8,16,1024]$ prompt bank is registered as one AdamW parameter
while the router and policy remain frozen. Current proxy gradients are sparse
in the selected prompt slices, but decay and historical moments are maintained
for the complete prompt bank. AGV therefore snapshots, commits, and rolls back both
the complete prompt bank and optimizer state.

\paragraph{Test-time optimization settings.}
Offline TTT performs suite-wide prompt-only AdamW optimization with learning
rate $10^{-5}$ and batch size 64 for 500 steps on WidowX and 1000 steps on
Google Robot, after which the adapted prompt is fixed during evaluation. Dense
Online TTT uses batch size 1 and takes one AdamW step with learning rate
$10^{-7}$ whenever a valid WPI pair arrives. Each AGV proposal consists of one
shadow AdamW step with learning rate $10^{-7}$, $\beta=(0.9,0.999)$, and weight
decay $0.01$; the candidate prompt and optimizer state are committed only after
future validation. These settings update latent prompts only; the router and
policy backbone remain frozen.

\section{AGV-TTT Implementation Details}
\label{app:agv_details}

\paragraph{Robust event gate.}
The change score in Eq.~\eqref{eq:event_change_compact} is normalized against a
rolling history $\mathcal H_t^c$:
\begin{equation}
    r_t^c=
    \frac{\Delta_t^c-\operatorname{median}(\mathcal H_t^c)}
    {\max(1.4826\operatorname{MAD}(\mathcal H_t^c),\epsilon)}.
    \label{eq:app_robust_score}
\end{equation}
Here $\operatorname{MAD}$ denotes median absolute deviation and $\epsilon>0$
is a numerical-stability constant.
An event requires both an absolute and a robust threshold in either channel.
Warm-up, cooldown, and a single-pending-candidate guard suppress startup noise
and overlapping proposals.

\paragraph{Focused token construction.}
Predictive-latent attention is normalized per layer and flow step and then
aggregated over selected late DiT layers. Let $\mathcal S$ denote the selected
flow-step--layer pairs and $a^{\mathrm{jepa}}_{t,u,l,n}$ the predictive-latent
attention weight for visual token $n$ among $N$ visual tokens:
\begin{equation}
\begin{aligned}
    A_n&=\frac{1}{|\mathcal S|}\sum_{(u,l)\in\mathcal S}
    \frac{a^{\mathrm{jepa}}_{t,u,l,n}}
         {\sum_{j=1}^{N}a^{\mathrm{jepa}}_{t,u,l,j}},\\
    F_n&=\frac{A_n}{\sum_{j\in\mathcal K_t}A_j},
    \qquad n\in\mathcal K_t.
\end{aligned}
    \label{eq:app_focus_weights}
\end{equation}
Here $A_n$ is the aggregated event-frame attention and $F_n$ is its normalized
weight over $\mathcal K_t$, corresponding to $F_{t,n}$ in the main text.
The main experiments use the top 32 tokens. Single-prompt WPI and WPI + MoLP
aggregate the last four and eight layers, respectively. The indices and
weights are created from the live event-frame attention, detached, and reused
for proposal training and all validation pairs.

\paragraph{Paired future validation.}
For each validation pair, old and candidate prompts receive identical present
and future latents, focus weights, Gaussian action noise, and flow time. CPU
and CUDA random states are reset before each branch and restored afterward.
Since no expert action is available, both branches use the same zero
action anchor; only future-latent prediction errors are compared. For
$x\in\{\mathrm{focus},\mathrm{global}\}$,
\begin{equation}
    R_x=
    \frac{\overline{\mathcal L}^{\mathrm{old}}_x-
          \overline{\mathcal L}^{\mathrm{cand}}_x}
    {\max(|\overline{\mathcal L}^{\mathrm{old}}_x|,\epsilon)}.
    \label{eq:app_relative_gain}
\end{equation}
The overline denotes the mean over the collected validation pairs, and the same
numerical-stability constant $\epsilon$ is used in the denominator.
The WPI setting uses $H_v=4$, focus weight $\lambda_f=1$, zero improvement
margin, and zero global-degradation tolerance. Finite-loss, parameter, and step
checks are applied before commit. A candidate that cannot finish validation
within an episode is discarded.

\begin{algorithm}[H]
\caption{WPI-AGV-TTT at deployment time}
\label{alg:wpi_agv_ttt}
\footnotesize
\begin{algorithmic}[1]
\STATE Initialize live prompt/optimizer $(\mathcal P,m)$ and no candidate
\FOR{each control step $t$}
    \IF{a candidate is pending and a new WPI pair is available}
        \STATE Evaluate old and candidate with paired randomness
        \IF{$H_v$ validation pairs have been collected}
            \STATE Commit iff Eq.~\eqref{eq:acceptance_compact} holds; otherwise retain live state
            \STATE Clear the candidate
        \ENDIF
    \ENDIF
    \STATE Generate $a_t$ with the live prompt and capture attention
    \IF{an event is detected and no candidate is pending}
        \STATE Freeze the attention-derived focus specification
        \STATE Take one shadow AdamW step on $\mathcal L_{\mathrm{focus}}$
        \STATE Restore live state and mark the candidate as pending
    \ENDIF
    \STATE Execute the already generated action $a_t$
\ENDFOR
\end{algorithmic}
\end{algorithm}

{\footnotesize
\noindent\textbf{State-grounded AGV.}
For the state-grounded baseline, AGV uses only action-to-VLM attention, the
state-grounding proxy, and a three-pair majority validation. Its acceptance
margin is $10^{-6}$, whereas WPI uses zero. This variant tests whether the
proposal-validation-commit abstraction transfers to another proxy; the full
VANE method is the dual-attention WPI instance.
\par}

\end{document}